\documentclass[final]{cvpr}

\usepackage{times}
\usepackage{epsfig}
\usepackage{graphicx}
\usepackage{amsmath}
\usepackage{amssymb}
\usepackage{booktabs,caption,siunitx}
\usepackage[dvipsnames]{xcolor}  
\usepackage{wrapfig}
\usepackage{subcaption}
\usepackage{floatrow}
\usepackage{caption}

\newfloatcommand{capbtabbox}{table}[][\FBwidth]

\makeatletter
\@namedef{ver@everyshi.sty}{}
\makeatother
\usepackage{tikz}
\usepackage{pgfplots, pgfplotstable}
\pgfplotsset{compat=newest}

\usepackage[inline, shortlabels]{enumitem}
\setlist[itemize]{leftmargin=*, noitemsep, topsep=0pt}
\setlist[enumerate,1]{leftmargin=*, noitemsep, topsep=0pt, label={\bf (\roman*)}}

\newcommand{\bluecheck}{{\color{Blue}\checkmark}}
\usepackage{pifont}
\newcommand{\xmark}{\color{BrickRed}\ding{55}}%

\usepackage[pagebackref=true,breaklinks=true,colorlinks,linkcolor=PineGreen,citecolor=PineGreen,bookmarks=false]{hyperref}

\newcommand{\beginsupplement}{%
        \setcounter{table}{0}
        \renewcommand{\thetable}{S\arabic{table}}%
        \setcounter{figure}{0}
        \renewcommand{\thefigure}{S\arabic{figure}}%
        }

\def\cvprPaperID{4172} 
\def\confYear{CVPR 2021}

\newcommand{\hts}{{How2Sign\xspace}}

\begin{document}

\title{How2Sign: A Large-scale Multimodal Dataset \\ for Continuous American Sign Language}

\author{Amanda Duarte$^{1,2*}$
\and
Shruti Palaskar$^4$
\and
Lucas Ventura$^1$
\and
Deepti Ghadiyaram$^5$
\and
Kenneth DeHaan$^6$
\and
Florian Metze$^{4,5}$
\and
Jordi Torres$^{1,2}$
\and
Xavier Giro-i-Nieto$^{1,2,3*}$
\and
\small$^1$\emph{Universitat Polit\`{e}cnica de Catalunya} \quad $^2$\emph{Barcelona Supercomputing Center} \quad 
\small$^3$\emph{Institut de Robòtica i Informàtica Industrial, CSIC-UPC}\\
\small$^4$\emph{Carnegie Mellon University} \quad $^5$\emph{Facebook AI} \quad $^6$\emph{Gallaudet University}
}

\twocolumn[{%
\renewcommand\twocolumn[1][]{#1}%

\maketitle

\begin{center}
    \centering
    \includegraphics[width=.88\textwidth]{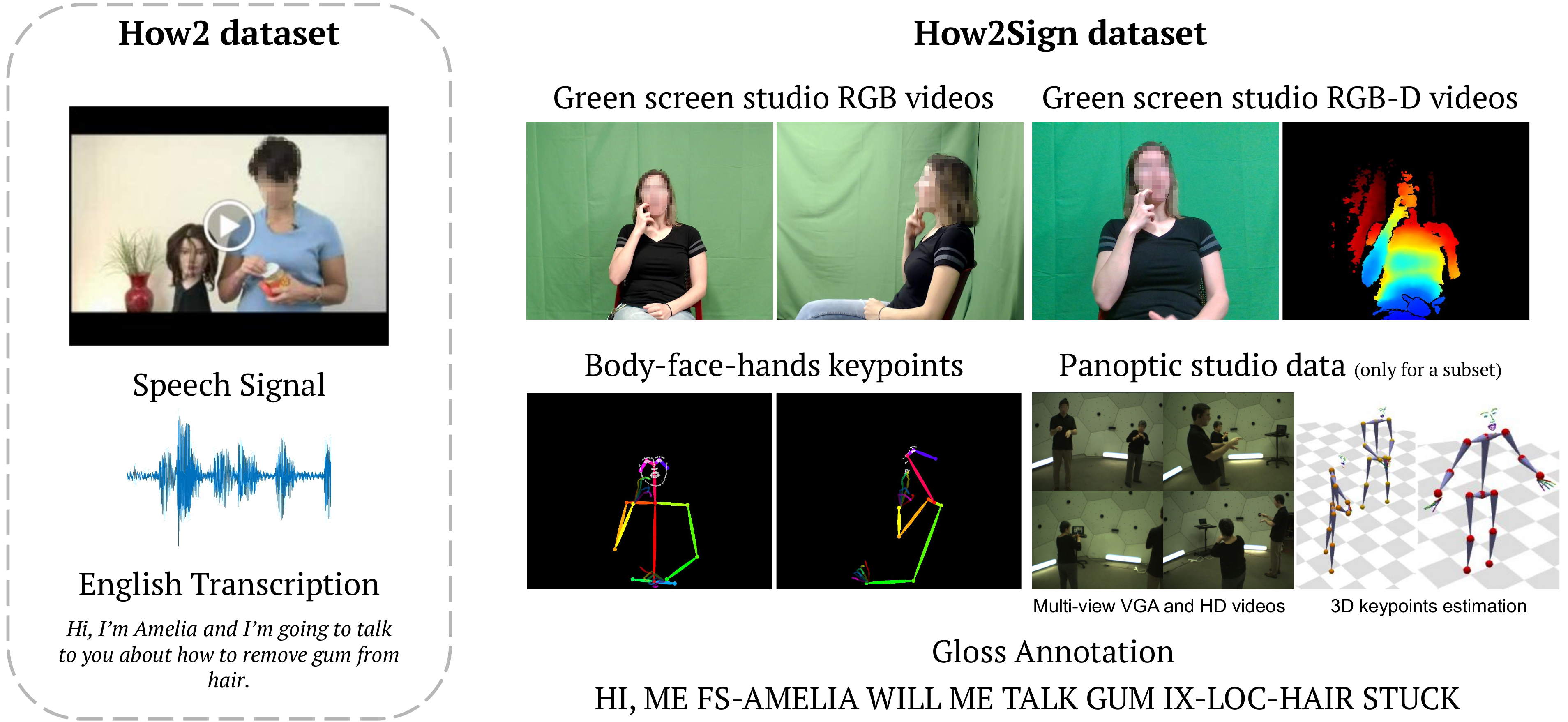}
    \captionof{figure}{The \textbf{How2Sign} dataset consists of over 80 hours of multiview sign language videos and aligned modalities.}
    \label{fig:teaser}
\end{center}
}]%

\begin{abstract}
One of the factors that have hindered progress in the areas of sign language recognition, translation, and production is the absence of large annotated datasets. 
Towards this end, we introduce \hts{}, a multimodal and multiview continuous American Sign Language (ASL) dataset, consisting of a parallel corpus of more than 80 hours of sign language videos and a set of corresponding modalities including speech, English transcripts, and depth. A three-hour subset was further recorded in the Panoptic studio
enabling detailed 3D pose estimation. 
To evaluate the potential of \hts{} for real-world impact, we conduct a study with ASL signers and show that synthesized videos using our dataset can indeed be understood. The study further gives insights on challenges that computer vision should address in order to make progress in this field.

\noindent Dataset website: \url{http://how2sign.github.io/}

\noindent\small*Corresponding authors: \{amanda.duarte,xavier.giro\}@upc.edu
\end{abstract}

\vspace{-20pt}
\section{Introduction}
\label{sec:introduction}

Sign languages (SL) are the primary means of communication for an estimated 466 million deaf
\footnote{We follow the recognized convention of using the upper-cased word Deaf which refers to the culture and describes members of the community of sign language users and the lower-cased word deaf describes the hearing status\cite{woodward1972implications}.} 
or hard-of-hearing people worldwide~\cite{deafstats}. 
Like any other natural language, sign languages are consistently evolving and have structure directed by a set of linguistic rules~\cite{sl-structure}. 
They differ from spoken languages and do not have standard written forms, \eg American Sign Language (ASL) is not a sign form of English. 
Although sign languages are used by millions of people everyday to communicate, the vast majority of communications technologies nowadays are designed to support spoken or written language, but not sign languages. At the same time, most hearing people do not know a sign language; as a result, many communication barriers exist for deaf sign language users~\cite{bragg2019sign, sl-barriers, goldmann1992overcoming}. 

Promising recent works in sign language processing\footnote{For brevity, we follow~\cite{bragg2019sign} and use the term \emph{sign language processing} to refer to the set of sign language recognition, translation and production tasks.}
~\cite{SLTranslation,progressive_transformers,Text2Sign, Czech_Sign_Language, WordsAreOurGlosses,korean_SL} have shown that modern computer vision and machine learning architectures can help break down these barriers for sign language users.
Improving such models could make technologies that are primarily designed for non-sign language users, \eg voice-activated services, text-based systems, spoken-media based content, \etc, more accessible to the Deaf community.   
Other possibilities include automatic transcription of signed content, which would help facilitating the communication between sign and non-sign language users, as well as real-time interpreting when human interpreters are not available, and many other educational tools and applications~\cite{bragg2019sign}. 

However, training such models requires large amounts of data. 
The availability of public large-scale datasets suitable for machine learning is very limited, especially when it comes to \emph{continuous sign language} datasets, \ie, where the data needs to be segmented and annotated at the sentence level. Currently, there is no ASL dataset large enough to be used with recent deep learning approaches.

In order to instigate the advance in the area of research that involves sign language processing, in this paper we introduce the \emph{\hts{}} dataset.
\hts{} is a large-scale collection of multimodal and multiview sign language videos in American Sign Language (ASL) for over 2500 instructional videos selected from the existing How2 dataset~\cite{How2}. Figure~\ref{fig:teaser} shows samples of the data contained in the dataset. Working in close collaboration with native ASL signers and professional interpreters, we collected more than \emph{80 hours} of multi-view and multimodal (recorded with multiple RGB and a depth sensor) ASL videos, and corresponding gloss annotations~\cite{gloss}. In addition, a three-hour subset was further recorded at the Panoptic studio~\cite{panoptic}, a geodesic dome setup equipped with hundreds of cameras and sensors, which enables detailed 3D reconstruction and pose estimation. This subset paves the way for vision systems to understand the 3D geometry of sign language.

Our contributions can be summarized as follows:
a) We present \hts{}, a large-scale multimodal and multiview continuous American Sign Language dataset that consists of more than \emph{80 hours of American Sign Language videos}, with sentence-level alignment for more than 35k sentences. It features a vocabulary of 16k English words that represent more than two thousand instructional videos from a broad range of categories;
b) Our dataset comes with a rich set of annotations including gloss, category labels, as well automatically extracted 2D keypoints for more than 6M frames. What is more, a subset of the dataset was re-recorded in the Panoptic studio with more than 500 cameras that enabled high quality 3D keypoints estimation for around 3 hours of videos;
c) We conduct a study with ASL signers that showed that videos generated using our dataset can be understood to a certain extent, and at the same time gave insights on challenges that the research community can address in this field.

\label{sec:intro}
\begin{table*}[t]
\centering
\caption{\textbf{Summary of publicly available continuous sign language datasets}. 
To the best of our knowledge, \hts{} is the largest publicly available Sign Language dataset across languages in terms of vocabulary, as well as the largest American Sign Language (ASL) dataset in terms of video duration.  We also see that \hts{} is the dataset with the most parallel modalities. A detailed explanation of each modality can be found in the subsection~\ref{subsec:content}
}
\label{tab:related_datasets}
\renewcommand{\arraystretch}{1.2}
\resizebox{\textwidth}{!}{\begin{tabular}{lccccccccccc} 
    \hline
        \textbf{Name}                            & \textbf{Language} & \textbf{Vocab.} & \textbf{Duration (h)} & \textbf{Signers} &  \multicolumn{6}{c}{\textbf{Modalities}} \\
                                                 &                   &           &               & &\textbf{Multiview} & \textbf{Transcription} & \textbf{Gloss} & \textbf{Pose} & \textbf{Depth} & \textbf{Speech} \\
    \hline                          
        Video-Based CSL~\cite{video-based}       & CSL               & 178       & 100           & 50 &\xmark & \bluecheck & \xmark     & \bluecheck     & \bluecheck   & \xmark \\
        SIGNUM~\cite{SIGNUM}                     & DGS               & 450       & 55            & 25 &\xmark     & \bluecheck & \bluecheck & \xmark     & \xmark   & \xmark \\
        RWTH-Phoenix-2014T~\cite{SLTranslation}  & DGS               & 3k        & 11            & 9 &\xmark     & \bluecheck & \bluecheck & \xmark     & \xmark   & \xmark \\
        Public DGS Corpus~\cite{DGS-Korpus}      & DGS               & --        & 50            & 327 &\bluecheck & \bluecheck & \bluecheck & \bluecheck & \xmark   & \xmark \\
        BSL Corpus~\cite{bsl-corpus}             & BSL               & 5k        & --            & 249 &\xmark     & \bluecheck & \bluecheck & \xmark     & \xmark   & \xmark \\ 
    \hline
        Boston104~\cite{Boston104}               & ASL               & 104       & 8.7 (min)     & 3 &\xmark     & \bluecheck & \bluecheck & \xmark     & \xmark     & \xmark \\
        NCSLGR~\cite{NCSLGR}                     & ASL               & 1.8k     & 5.3            & 4 &\bluecheck & \bluecheck & \bluecheck & \xmark     & \xmark     & \xmark  \\
    \hline
        \textbf{How2Sign (ours)}                 & ASL               & 16k       & 79            & 11 &\bluecheck & \bluecheck & \bluecheck & \bluecheck & \bluecheck & \bluecheck \\
    \hline
    
\end{tabular}}
\end{table*}

\section{Background and Related Work}
In this section we discuss some of the challenges that comes with sign languages that can be interesting to the computer vision community, as well as an overview of the current publicly available sign language datasets. 

\subsection{Sign Language}
Sign languages are visual languages that use two types of features to convey information: \emph{manual} that includes handshape, palm orientation, movement and location and; \emph{non-manual markers} that are movement of the head (nod/shake/tilt), mouth (mouthing), eyebrows, cheeks, facial grammar (or facial expressions) and eye gaze~\cite{ASLstructure}.
All these features need to be taken into account while recognizing, translating or generating signs in order to capture the complete meaning of the sign. This makes sign language processing a challenging set of tasks for computer vision.

When it comes to continuous sign language, a simple concatenation of isolated signs is not enough to correctly recognize, translate or generate a complete sentence and neglects the underlying rich grammatical and linguistic structures of sign language that differ from spoken language. Besides the fact that the alignment between sign and spoken language sequences are usually unknown and non monotonic~\cite{SLTranslation}, the transitions between signs must also be taken into account.  
Usually, the beginning of a sign is modified depending on the previous sign, and the end of the same sign is modified depending on the following sign making them visually different in the isolated and continuous scenarios~\cite{sl-structure}. This phenomenon is called “co-articulation” and brings an extra challenge for tasks with continuous sign language~\cite{BSL-1K}.

\subsection{Sign Language datasets}

One of the most important factors that has hindered the progress of sign language processing research is the absence of large scale annotated datasets~\cite{bragg2019sign}.
Many existing sign language datasets contain isolated signs~\cite{ASL-Lex,ASLLVD,MS-ASL,WLASL,RVL-SLLL,signor}. Such data may be important for certain scenarios (\eg, creating a dictionary, or as a resource for those who are learning a sign language), but most real-world use cases of sign language processing involve natural conversational with complete sentences (\ie \emph{continuous sign language}).

A number of continuous sign language datasets have been collected over the years mainly for linguistic purposes. 
SIGNUM~\cite{SIGNUM} and the BSL Corpus~\cite{bsl-corpus} were recorded in controlled environments with a single RGB camera.  
Recent works in neural machine translation~\cite{camgoz2020sign} and production~\cite{progressive_transformers,saunders2020adversarial} have adopted~\textit{RWTH-Phoenix-2014T}~\cite{SLTranslation}, a dataset of German Sign Language (DGS) on the specific domain of weather forecast from a TV broadcast that features 9 signers.
The \textit{Public DGS Corpus}~\cite{DGS-Korpus} and the \textit{Video-Based CSL (Chinese Sign Language)}\cite{video-based} provide much larger video collections enriched with the body keypoint of the signers. In the case of \textit{Public DGS Corpus}, these are 2D poses estimated with OpenPose~\cite{openpose} and from different view points, while \textit{Video-Based CSL} provides 3D joints and depth information thanks to the recordings with a Kinect camera.
If we focus on American Sign Language (ASL), RWTH-BOSTON-104~\cite{Boston104} only contains 8.7 minutes of grayscale video, while NCSLGR \cite{NCSLGR} is larger but an order of magnitud smaller than How2Sign.
In terms of annotation, all datasets but \textit{Video-Based CSL} provide gloss annotations, that is, a text-based transcription of the signs that can serve as a proxy in translation tasks.

Table~\ref{tab:related_datasets} presents an overview of publicly available continuous sign language datasets ordered by vocabulary size\footnote{An extended overview of related datasets can be found at: \url{https://how2sign.github.io/related_datasets.html}}. 
An important factor for the lack of large-scale datasets is that the collection and annotation of continuous sign language data is a laborious and expensive task. It requires linguistic experts working together with a native speaker, e.g a Deaf person.
RWTH-Phoenix-2014T~\cite{SLTranslation} is one of the few datasets that are publicly available and has been used for training deep neural networks. A recent re-alignment in the annotations also allows studying sign language translation.
However, their videos cover just 11 hours of data from weather broadcasts, and are restricted to one domain.

In summary, the current publicly available datasets are constrained by one or more of the following: (i) limited vocabulary size, (ii) short video or total duration and (iii) limited domain. 
The \hts{} dataset provides a considerably larger vocabulary than the existing ones, and it does so in the continuous sign language setting for a broader domain of discourse. It also is the first sign language dataset that contains speech thanks to its alignment with the existing How2 dataset \cite{How2}.

\label{sec:related_work}
\section{The How2Sign dataset}
The \hts{} dataset consists of a parallel corpus of speech and transcriptions of instructional videos and their corresponding American Sign Language (ASL) translation videos and annotations. A total of \emph{80 hours} of multiview American Sign Language videos were collected, as well as gloss annotations~\cite{gloss} and a coarse video categorization.

\noindent\textbf{Source language.}
The instructional videos translated into ASL come from the existing \emph{How2 dataset}~\cite{How2}, a publicly available multimodal dataset for vision, speech and natural language understanding, with utterance-level time alignments between the speech and the ground-truth English transcription. Following the same splits from the \emph{How2-300h} dataset, we selected a 60-hour subset from the training set and the complete validation and test sets to be recorded.

\subsection{Sign language video recordings}
\label{sec:recordings}
\begin{figure*}[t!]
    \centering
    \begin{subfigure}[t]{.49\textwidth}
        \centering
        \includegraphics[width=0.85\textwidth]{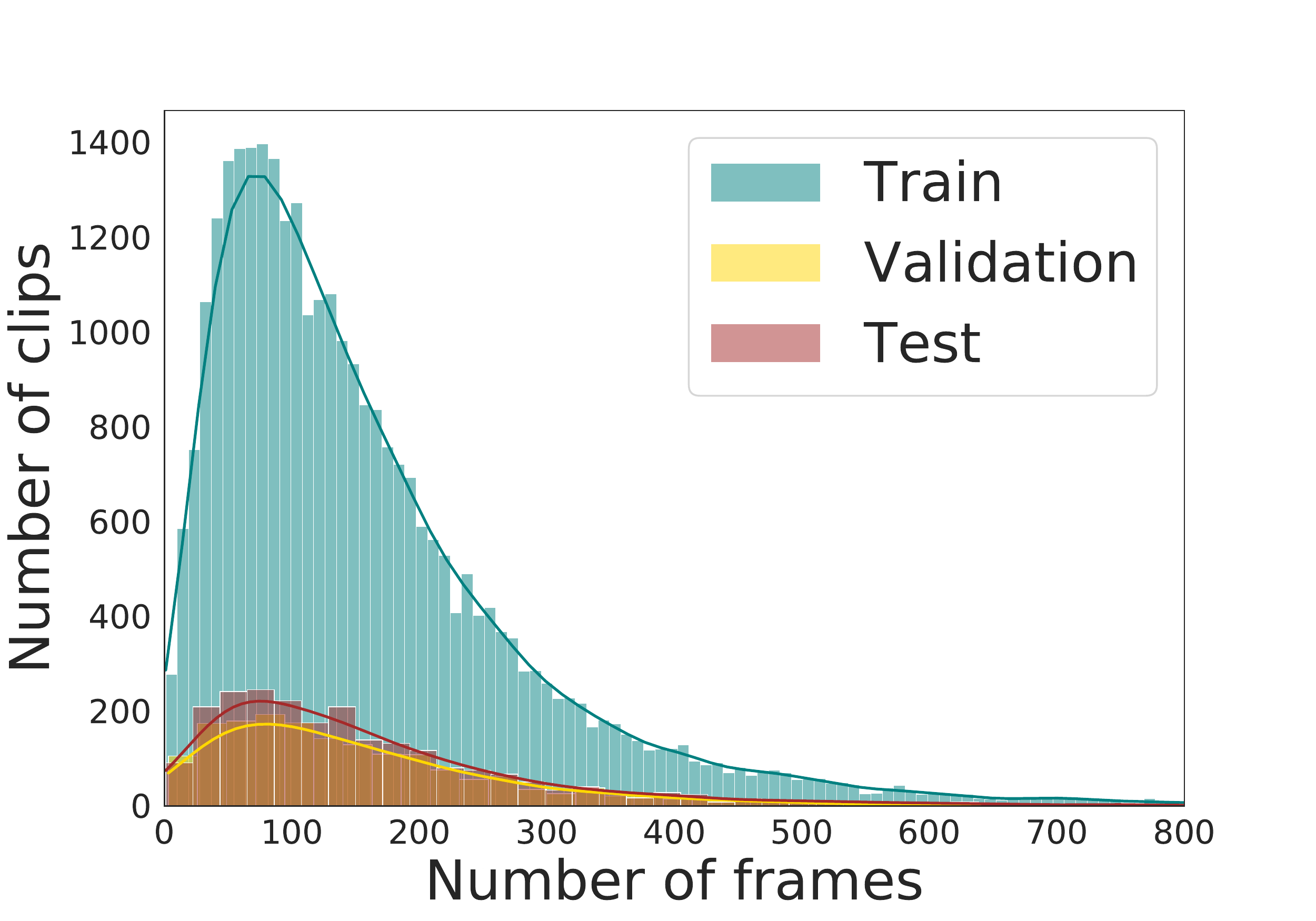}
    \end{subfigure}%
    \begin{subfigure}[t]{0.49\textwidth}
        \centering
        \includegraphics[width=0.85\textwidth]{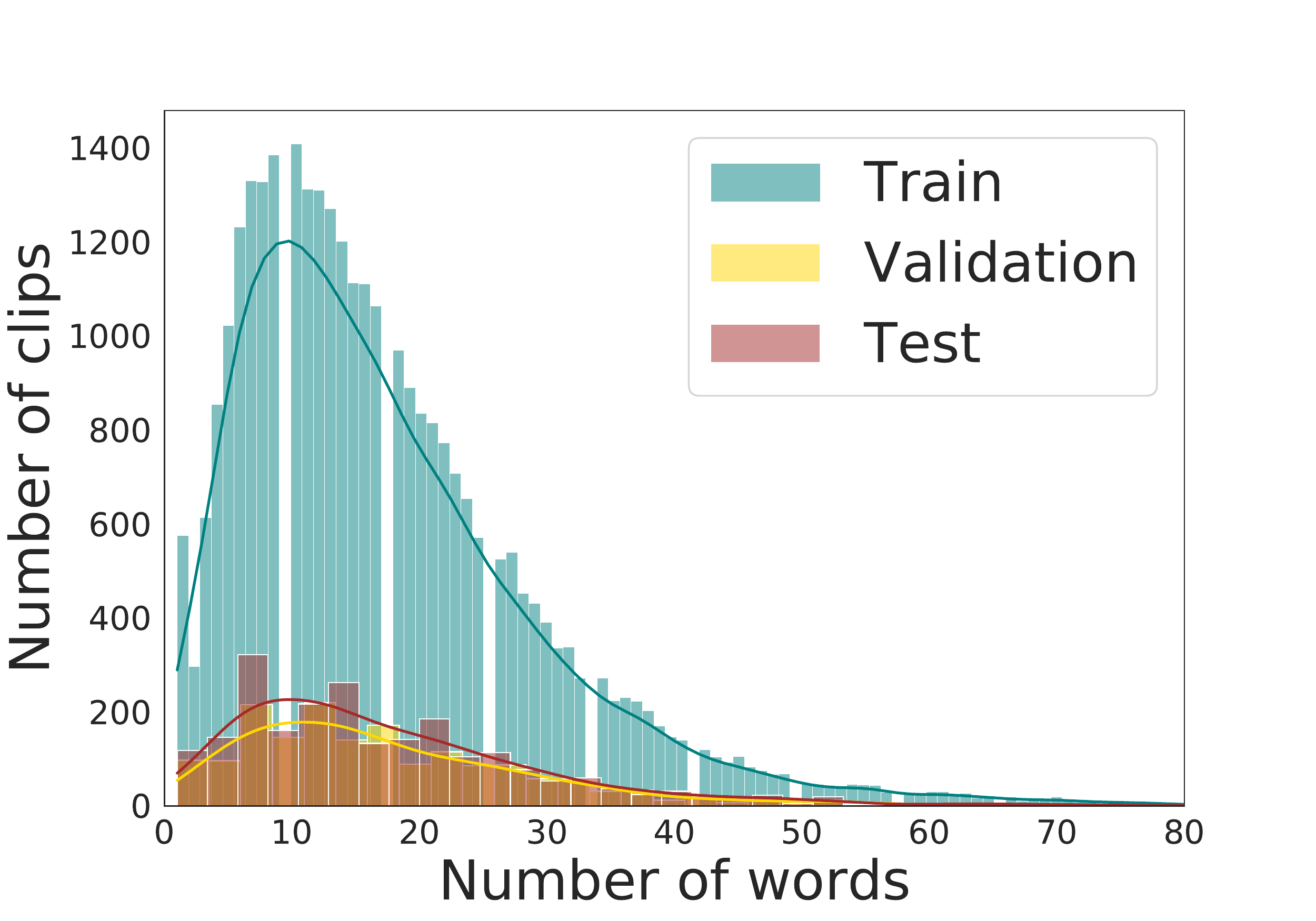}
    \end{subfigure}
    \caption{Distribution of the number of frames (left) and words (right) over sentence-level clips.}
    \label{fig:plots}
\end{figure*}

\noindent\textbf{Signers.} In total, 11 people appear in the sign language videos of the \hts{} dataset; we refer to them as \emph{signers}. Of the 11 signers, 5 self-identified as hearing, 4 as Deaf and 2 as hard-of-hearing. The signers that were hearing were either professional ASL interpreters (4) or ASL fluent. 

\noindent\textbf{Recording pipeline.}
The signer would first watch the video with the transcript as subtitles in order to become familiar with the overall content; this enables them to perform a richer translation. ASL translation videos were then recorded, while the signer was watching the video with subtitles, and at a slightly slower-than-normal (0.75) speed. For each hour of video recorded, the preparation, recording and video review took approximately 3 hours on average. 

All recordings were performed in a supervised setting in two different locations: at the \emph{Green Screen studio} and at the \emph{Panoptic studio}, both presented below. We recorded the complete 80 hours of the dataset in the green screen studio. We then chose a small subset of videos (approx. 3 hours) from the validation and test splits and recorded them again in the Panoptic studio. After recording, we trimmed all sign language videos and divided them in \emph{sentence-level clips}, each annotated with a corresponding English transcript, and the modalities presented in Section~\ref{subsec:content}. 

\noindent\textbf{Green screen studio.}
The \emph{Green Screen studio} was equipped with a depth and a high definition (HD) camera placed in a frontal view of the participant, and another HD camera placed at a lateral view. All three cameras recorded videos at 1280x720 resolution, at 30 fps. Samples of data recorded in this studio are shown in the top row of Figure \ref{fig:teaser}.

\noindent\textbf{Panoptic studio.}
The \emph{Panoptic studio}~\cite{panoptic} is a system equipped with 480 VGA cameras, 30 HD cameras and 10 RGB-D sensors, all synchronized. All cameras are mounted over the surface of a geodesic dome\footnote{\url{http://www.cs.cmu.edu/~hanbyulj/panoptic-studio/}}, providing redundancy for weak perceptual processes (such as pose detection) and robustness to occlusion. In addition to the multiview VGA and HD videos, the recording system can further estimate high quality 3D keypoints of the interpreters, also included in How2Sign.
Samples of data recorded in this studio are shown on the bottom-right of Figure \ref{fig:teaser}.

\subsection{Dataset Modalities}
\label{subsec:content}

The modalities enumerated in the columns of Table~\ref{tab:related_datasets} are detailed in this section.
Apart from the English translations and speech modalities that were already available from the How2~\cite{How2} dataset, all other modalities were either collected or automatically extracted. To the best of our knowledge, \hts{} is the largest publicly available sign language dataset across languages in terms of vocabulary, as well as an order of magnitude larger than any other ASL dataset in terms of video duration. We see that \hts{} is also the dataset with the most parallel modalities, enabling multimodal learning. 

\noindent\textbf{Multiview.} All 80 hours of sign language videos were recorded from multiple angles. This allows the signs to be visible from multiple points of view, reducing occlusion and ambiguity, especially in the hands. Specifically, the sign language videos recorded in the Green Screen studio contain two different points of view, while the Panoptic studio recordings consist of recordings of more than 500 cameras allowing for a high quality estimation of 3D keypoints~\cite{panoptic}.

\noindent\textbf{Transcriptions.} The English translation modality originates from the subtitles track of How2 original videos. The transcriptions were provided by the uploader of the instructional video in form of text, that was loosely synced with the video's speech track. As subtitles are not necessarily fully aligned with the speech, transcriptions were time-aligned at the sentence-level as part of the How2 dataset~\cite{How2}. 

\noindent\textbf{Gloss} is used in linguistics to transcribe signs using spoken language words. It is generally written in capital letters and indicates what individual parts of each sign mean, including annotations that account for facial and body grammar. An example of gloss annotation is shown on the bottom right of Figure \ref{fig:teaser}.
It is important to note that gloss is not a true translation, it instead provides the appropriate spoken language morphemes that express the meaning of the signs in spoken language~\cite{gloss-informal,gloss}. Glosses do not indicate special hand-shape, hand movement/orientation, nor information that would allow the reader to determine how the sign is made, or what its exact meaning in a given context. They also do not indicate grammatical uses of facial expressions (for example, raising the eyebrows is used in yes/no questions). 
Gloss is the form of text that is closest to sign language and it has been used by a number of approaches as an intermediate representation for sign language processing~\cite{SLTranslation,progressive_transformers,saunders2020adversarial,WordsAreOurGlosses,korean_SL}.

\noindent\textbf{Pose information.}
Human pose information, \eg body, hand and face keypoints were extracted for all the recorded sign language videos in the full resolution -- 1280 x 720 pixels.
For the Green Screen studio data, the 2-dimensional (2D) pose information was automatically extracted using OpenPose~\cite{openpose}. In total, each pose consists of 25 body keypoints, 70 facial keypoints and 21 keypoints for each hand. We provide pose information for both frontal and side view of the Green Screen studio data. 
A sample of the pose information extracted can be seen on the bottom row in the left side of Figure~\ref{fig:teaser}. For the Panoptic studio data, we provide high quality 3-dimensional (3D) pose information estimated by the Panoptic studio internal software~\cite{panoptic} that can be used as ground-truth for a number of 3D vision tasks.

\noindent\textbf{Depth data.}
For the Green Screen studio data, the sign language videos were also recorded using a Depth sensor (Creative BlasterX Senz3D) from the frontal viewpoint. The sensor has high precision facial and gesture recognition algorithms embedded and is able to focus on the hands and face, the most important human parts for sign language. 

\noindent\textbf{Speech.}
The speech track comes from the instructional videos as part of the How2 dataset~\cite{How2}.

\subsection{Collected Annotations}
\label{subsec:annotations}

Beyond the video recordings and automatically extracted pose information, we further collected a number of manual annotations for the sign language videos.

\noindent\textbf{Gloss and sentence boundaries.}
We collected gloss annotations by employing ASL linguists. The annotations were collected using ELAN~\cite{ELAN}, an annotation software for audio and video recordings, specifically enhanced for sign language annotations. 
Information in ELAN is represented in tiers which are time-aligned to the video files, giving us the start and end boundaries of each sentence and producing what we call the sentence boundaries. 
The gloss annotation took in average one hour per 90 seconds of video.

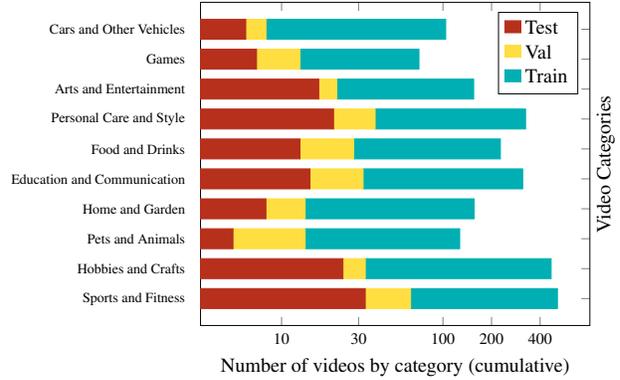
\begin{figure}
    \centering
    \resizebox{\columnwidth}{!}{
    \begin{tikzpicture}
    	\begin{semilogxaxis}[xbar stacked,
            ytick = {1,2,3,4,5,6,7,8,9,10},
            yticklabels = {Sports and Fitness, Hobbies and Crafts, Pets and Animals, Home and Garden, Education and Communication, Food and Drinks, Personal Care and Style, Arts and Entertainment, Games, Cars and Other Vehicles},
            ylabel={Video Categories},
            ylabel style = {font=\footnotesize},
            xlabel={Number of videos by category (cumulative)},
            ylabel near ticks, yticklabel pos=right,
            ylabel style={xshift=0.cm},
            xtick = {10,30, 100,200,400},
            xminorticks=true,
            minor x tick num=10,
            xticklabels = {10,30,100,200,400},
            yticklabel style = {font=\scriptsize, xshift=-7cm, anchor=east, rotate=0},
            xticklabel style = {font=\footnotesize,yshift=-0.1ex},
            legend cell align=left,
            legend style={legend pos=north east,font=\normalsize}],
            
    	]
    
    	\addplot[BrickRed, fill=BrickRed] coordinates
    		{
                (33, 1)
                (24, 2)
                (5, 3)
                (8, 4)
                (15, 5)
                (13, 6)
                (21, 7)
                (17, 8)
                (7, 9)
                (6, 10)
            }; \addlegendentry{Test};
    		
    	\addplot[Goldenrod, fill=Goldenrod] coordinates
    		{
    		(30,1)
            (9,2)
            (9,3)
            (6,4)
            (17,5)
            (15,6)
            (17,7)
            (5,8)
            (6,9)
            (2,10)
            }; \addlegendentry{Val};

        \addplot[TealBlue, fill=TealBlue] coordinates
    		{
    		(449, 1)
            (435, 2)
            (113, 3)
            (142, 4)
            (280, 5)
            (199, 6)
            (287, 7)
            (133, 8)
            (58, 9)
            (96, 10)
    		}; \addlegendentry{Train};
    		
    	\end{semilogxaxis}
    \end{tikzpicture}
    }
    \caption{Cummulative number of videos per category.}
    \label{fig:video_categories}
\end{figure}

\noindent\textbf{Video Categories.}
Although the How2 dataset provides automatically extracted ``topics'' for all videos using Latent Dirichlet Allocation~\cite{LDA}, we found that the automatic annotations were in general very noisy and not properly characterizing the selected videos. 
In order to better categorize the videos, we manually selected 10 categories\footnote{The categories are: Personal Care and Style, Games, Arts and Entertainment, Hobbies and Crafts, Cars and Other, Vehicles, Sports and Fitness, Education and Communication, Food and Drinks, Home and Garden and Pets and Animals.} from the instructional website Wikihow\footnote{https://www.wikihow.com/Special:CategoryListing} and manually classified each \hts{} video in a single category. The distribution of videos across the ten categories can be seen in Figure~\ref{fig:video_categories}.

\subsection{Dataset statistics}

\begin{table*}[t]
\begin{center}
\resizebox{.8\textwidth}{!}{%

    \begin{tabular}{l lll| l | ll | l} 
     \toprule
                              & \multicolumn{4}{c}{\textbf{Green screen studio}}                   & \multicolumn{3}{c}{\textbf{Panoptic studio}} \\
                              & \textbf{train} & \textbf{val} & \textbf{test} & \textbf{Total}     & \textbf{val}  & \textbf{test} & \textbf{Total}              \\ 
     \midrule
        How2~\cite{How2} videos                   & 2,192         & 115           & 149      &  2,456   & 48             & 76         & 124         \\
        Sign language Videos                      & 2,213         & 132           & 184      &  2,529   & 48             & 76         & 124         \\
        Sign language video Duration (h)          & 69.62         & 3.91          & 5.59     &  79.12   & 1.14           & 1.82       & 2.96        \\
        Number of frames (per view)               & 6.3M          & 362,319       & 521,219  &  7.2M    & 123,120        & 196,560    & 319,680     \\
        Number of clips                           & 31,128        & 1,741         & 2,322    &  35,191  & 642            & 940        & 1,582        \\
    \midrule
        Camera views per SL video & \multicolumn{4}{c|}{3 HD + 1 RGB-D}       & \multicolumn{3}{c}{480 VGA + 30 HD + 10 RGB-D} \\ 
    \midrule
        
        Sentences                                 & 31,128        & 1,741         & 2,322        & 35,191   & 642            & 940  & 1,582  \\
        Vocabulary size                           & 15,686        & 3,218         & 3,670        &          & 1807           & 2360 & 3260    \\
        Out-of-vocabulary                         & –             & 413           & 510          &          &                &      &                       \\ 
        Number of signers                         & 8             & 5             & 6            & 9        & 3              & 5    & 6              \\
        Signers not in train set                  & –             & 0             & 1            &          & 2              & 2    &              \\
        \midrule
                                & \multicolumn{4}{c|}{\underline{\textit{2D keypoints}} }&  \multicolumn{3}{c}{ \underline{\textit{3D keypoints}}}  \vspace{2pt} \\
        Body pose               & \multicolumn{3}{c|}{25 }                               &     & \multicolumn{3}{c}{25 }   \\
        Facial landmarks        & \multicolumn{3}{c|}{70 }                               & 137 & \multicolumn{3}{c}{70 }   \\
        Hand pose (two hands)   & \multicolumn{3}{c|}{21 + 21 }                          &     & \multicolumn{3}{c}{21 + 21 } \\
     \bottomrule
    
    \end{tabular}
    }%
\vspace{-10pt}
  \caption{Statistics of the \textbf{\hts{}} dataset. Some of the videos were recorded more than once by a different signer in the Green screen studio (see second row vs. first row). ASL videos recorded were split into sentence-level \emph{clips}. Each clip has on average 162 frames (5.4 seconds) and 17 words.}%
  \label{tab:data_stats}
\end{center}
\end{table*}

In Table~\ref{tab:data_stats} we show detailed statistics of the \hts{} dataset.
A total of 2,456 videos from the How2~\cite{How2} were used to record the sign language videos. Some of the videos were recorded more than once by a different signer in the Green screen studio -- 21 videos from the training set, 17 videos from the validation set and 35 videos from the test set. All the recorded Videos were split into sentence-level clips. Each clip has on average 162 frames (5.4 seconds) and 17 words. The distribution of frames (right) and words (left) over all the clips for the 3 splits of the dataset can be seen in Figure~\ref{fig:plots}.
The collected corpus covers \emph{more than 35k sentences} with an English vocabulary of more than 16k words. Where approximately, 20\% of it is finger spelled.
The videos were recorded by 11 different signers distributed across the splits. 
The test set contains 26 duplicated videos that were recorded by a signer that is not present in the training set; this subset of 26 videos can be used for measuring \textit{generalization across different signers}.
In total, 9 signers participated in the Green Screen studio recordings, and 6 signers in the Panoptic studio recordings.
The bottom section of Table~\ref{tab:data_stats} refers to the automatically extracted human pose annotations. 

\subsection{Privacy, Bias and Ethical Considerations} 
In this section we discuss some metadata that we consider important for understanding the biases and generalization of the systems trained on our data.

\noindent\textbf{Privacy.} Since facial expressions are a crucial component for generating and/or translating Sign Language, it was not possible to avoid recordings that include the signer's face. To that end, all the research steps followed procedures approved by the Carnegie Mellon University Institutional Review Board including a Social \& Behavioral Research  training done by the first and second authors, and a consent form provided by the participants agreeing on being recorded and making their data publicly available for research purposes. It is important to note that this puts at risk the authenticity of the linguistic data collected, as signers may monitor their production more carefully than usual.

\noindent\textbf{Audiological status and language variety.} 
The majority of the participants identified American Sign Language and contact signing (Pidgin Sign English - PSE) as the main language used during the recordings. It is noteworthy that differences in audiological status are correlated with different language use. The Deaf were likely to identify ASL as the main language used in the recording process. In contrast, the hearing were likely to identify a mix of contact signing and ASL as the main language use in the recording process. More information about PSE and ASL can be found in~\cite{reilly1980american}.

\noindent\textbf{Geographic.} All participants were born and raised in the United States of America, and learned American Sign Language as their primary or second language at school time. 

\noindent\textbf{Signer variety.} Our dataset was recorded by signers with different body proportions. Six of them were self-identified male and five self-identified female. The dataset was collected across 65 days during 6 months which gives a variety of clothing and accessories used by the participants. 

\noindent\textbf{Data bias.} Our data does not contain large diversity in race/ethnicity, skin tone, background scenery, lighting conditions and camera quality.
\label{sec:dataset}
\section{Evaluating the potential of \hts{} for sign language tasks}
\label{sec:analysis}

The communication barrier between sign and non-sign language users may be reduced in the coming years thanks to the recent advances in neural machine translation and computer vision.
Recent works are making steps towards sign language production~\cite{progressive_transformers,Text2Sign, Czech_Sign_Language, WordsAreOurGlosses,saunders2020everybody} by automatically generating detailed human pose keypoints from spoken language, and translation~\cite{korean_SL}, \ie, using keypoints as input to generate text.

While keypoints can carry detailed human pose information and can be an alternative for reducing the computational bottleneck that is introduced when working with the actual video frames, no studies have been made so far on whether they are indeed useful when it comes to understanding sign language by its users.
In this section we present a study where we try to understand \textit{if and how well sign language users understand automatically generated sign language videos} that use keypoints from \hts{} as sign language representation. 
We run this study with four ASL speakers and record their understanding of the generated videos in terms of the category, translation into American English, and a final subjective rating about how understandable the videos were. 

\begin{figure*}
    \centering
    \includegraphics[width=.88\textwidth]{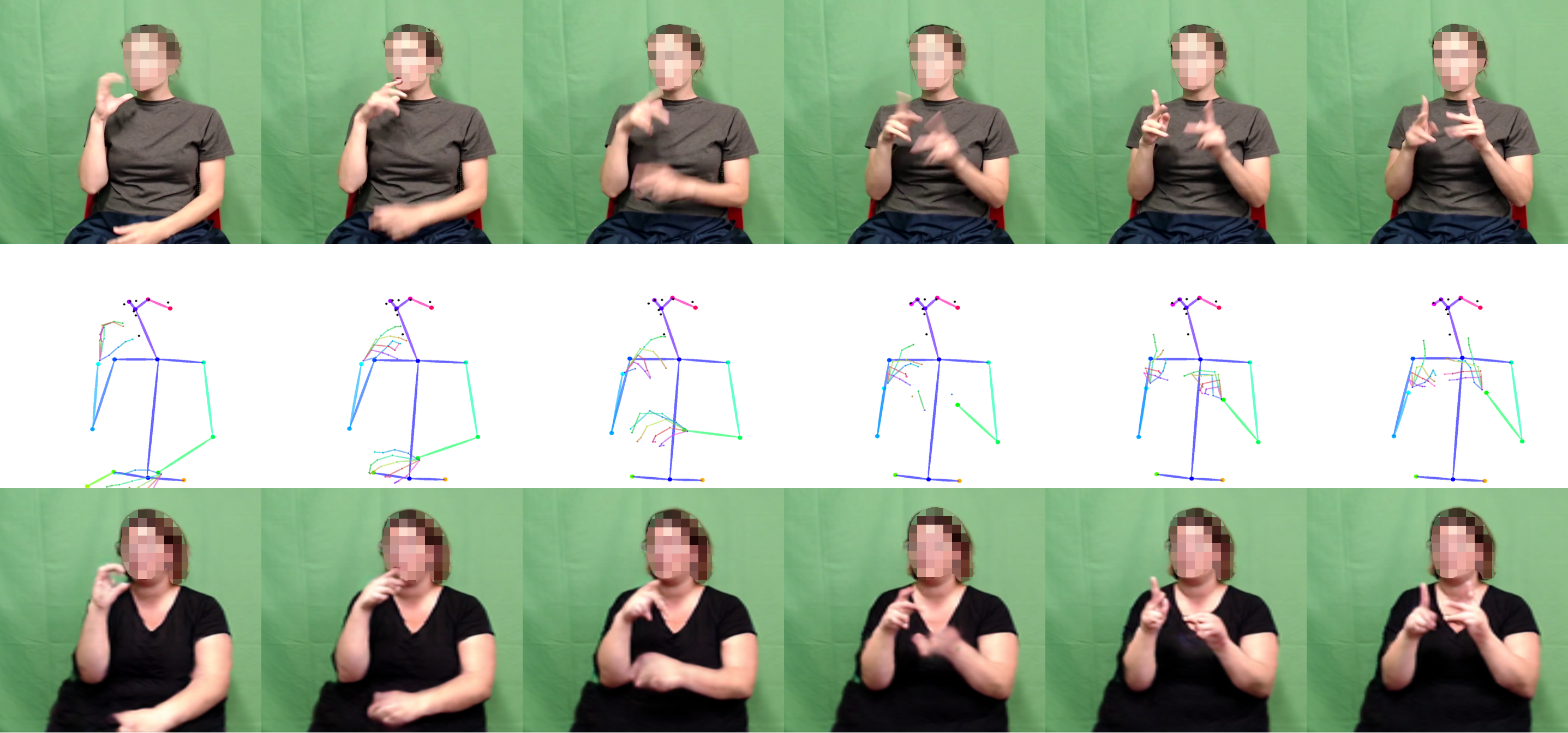}
    \caption{Sample of generated SL videos. Source video (top row) was used to automatically extract 2D keypoints (middle row) and generate frames of a video with a different identity (bottom row).}
    \label{fig:generated_video}
\end{figure*}

\subsection{Synthesizing sign language videos}
We experiment with two ways of generating sign language videos: 1) skeleton visualizations and 2) Generative Adversarial Network generated (GAN-generated) videos.
\noindent\textbf{Skeleton visualizations.} Given a set of estimated keypoints, one can visualize them as a wired skeleton connecting the modeled joints (see the middle row of Figure~\ref{fig:generated_video}). 
\noindent\textbf{GAN-generated videos.} Another option would be to go one step further and use generative models to synthesize videos on top of predicted keypoints. To generate the animated video of a signer given a set of keypoints, we use the motion transfer and synthesis approach called Everybody Dance Now (EDN)~\cite{EDN}. This model is based on Pix2PixHD~\cite{wang2018pix2pixHD}, but is further enhanced with a learned model of temporal coherence for better video and motion synthesis between adjacent frames by predicting two consecutive frames, as well as a separate module for high resolution face generation. 
It is worth noting that this approach models facial landmarks separately, something highly desirable in our case because they are one of the critical features for sign language understanding. 
The EDN model was trained on a subset of the \hts{} dataset that contains videos from \textit{two} participants. Specifically, keypoints extracted from videos of the first signer (top row in Figure~\ref{fig:generated_video}) were used to learn the model that generates realistic videos of the second signer (bottom row)\footnote{A sample of a generated video can be seen at: \url{https://youtu.be/wOxWUyXX6Ys}}.
The subset used consists of 28 hours of the training split. 

\subsubsection{Quantitative evaluation of the GAN-generated sign language videos.}

An approximate but automatic way of measuring the visual quality of the generated videos is by comparing the keypoints that can be reliably detected by OpenPose in the source and generated videos. We focus only on the 125 upper body keypoints which are visible in the \hts{} videos, and discard those from the legs. We use two metrics: a) the Percentage of Detected Keypoints (PDK), which corresponds to the fraction of keypoints from the source frame which were detected in the synthesized frame, and b) the Percentage of Correct Keypoints (PCK)~\cite{PCK}, which labels each detected keypoint as ``correct" if the distance to the keypoint in the original image is less than 20\% of the torso diameter in all keypoints and 10\% of the torso diameter for the hands.

In Table~\ref{tab:gan_video_eval} we present these metrics for different minimum confidence thresholds of the OpenPose (OP keypoint detectors).
We report results for all keypoints, as well as when restricting the evaluation only on the hand keypoints.
We see that although the repeatability of keypoints is high in general, the model fails to predict reliable keypoints for the hands. This limitation is especially relevant in sign language processing. 

\begin{table}[t]
\centering
\caption{Percentage of Detected Keypoints (PDK) and Percentage of Correct Keypoints (PCK) for all keypoints and just for the hands, when thresholding at different detection confidence scores of OpenPose (OP).} 
\label{tab:gan_video_eval}
    \resizebox{\columnwidth}{!}{
    \begin{tabular}{lccccccc}
    \toprule
                             & \multicolumn{3}{c|}{\textbf{PDK}}    & \multicolumn{3}{c}{\textbf{PCK}} \\
        OP confidence scores & 0 & 0.2 & \multicolumn{1}{c|}{0.5}   & 0 & 0.2 & 0.5 \\
    \midrule
        All keypoints   &  0.99   &  0.88  &  0.87  &   0.90  &  0.94   & 0.96  \\
        Hands           &  0.99   &  0.38  &  0.17  &   0.08  &  0.11   & 0.12   \\
    \bottomrule
\end{tabular}
     }
\end{table}


\subsection{Can ASL signers understand generated sign language videos?}
We evaluate the degree of understanding for both skeleton visualizations and the GAN-generated videos by showing 3-minute-long videos to four ASL signers. Two of them watched the skeletons visualizations, while the other two watched the GAN-generated videos. 
During the evaluation, each subject was asked to: a) classify six videos between the ten video categories (see subsection~\ref{subsec:content} for more information about the dataset categories); b) answer the question \textit{``How well could you understand the video?"} on the five-level scale ((1) Bad, (2) Poor, (3) Fair, (4) Good, (5) Excellent); and c) watch two clips from the previously seen video and translate them into American English. 
Results averaged over all subjects are presented in Table~\ref{tab:quantitative_results}. We report accuracy for the classification task, the Mean Opinion Score (MOS) for the five-scale question answers and BLEU~\cite{papineni2002bleu} scores for the American English translations. 
Qualitative results are shown in Table~\ref{tab:qualitative_results}.

Results show a preference towards the generated videos rather than the skeleton ones, as the former result in higher scores across all metrics. 
In terms of general understanding of the topic, the subjects were able to mostly classify the videos correctly with both types of visualizations. 

When it comes to finer grained understanding measured via the English translations, however, we can see from both Table~\ref{tab:quantitative_results} and Table~\ref{tab:qualitative_results} that neither skeletons nor GAN-generated videos are sufficient to convey important information needed from ASL signers to completely understand the sign language sentences.
We hypothesize that current human pose estimation methods such as~\cite{openpose} are still not mature enough when it comes to estimate fast movements of the hands. We observed that due to the nature of sign language and the fast movements of the signers' hands, OpenPose lacks precision in those cases which can make the visualizations incomplete, harming the understanding of some important parts of sign language. 

\begin{table}[t]
\centering

\caption{Comparison between generated skeletons and GAN videos in terms of classification (Accuracy), mean opinion score (MOS) and translation (BLEU)~\cite{papineni2002bleu}.} 
\label{tab:quantitative_results}
\resizebox{\columnwidth}{!}{
    \begin{tabular}{lc|c|cccc}
    \toprule
        & \textbf{Acc.} & \textbf{MOS} & \textbf{BLEU-1} &  \textbf{BLEU-2} &  \textbf{BLEU-3} & \textbf{BLEU-4} \\
    \midrule
        Skeleton & 83.3 \%            &    2.50       &  10.90          &       3.02    &      1.87     &       1.25   \\
        GAN-generated        & \textbf{91.6 \%}  & \textbf{2.58} &  \textbf{12.38} & \textbf{6.71} & \textbf{3.32} & \textbf{1.89}  \\ 
    \bottomrule
        \end{tabular}}
\end{table}

\begin{table}[t]
\caption{Ground-truth (GT) and collected translations for two clips of the ``Food and Drink" category. All subjects were able to correctly classify the category.}
\label{tab:qualitative_results}
\resizebox{\columnwidth}{!}{
    \begin{tabular}{ll}
        \toprule
            GT     &   \textbf{I'm not going to use a lot, I'm going to use very very little.} \\ 
        \midrule 
            Skeleton        &   That is not too much                                                    \\  
                            &   don't use much, use a little bit                                        \\ \cline{2-2}
            EDN             &   Don't use a lot, use a little                                           \\
                            &   dont use lot use little bit                                             \\ 
        \midrule
            GT     &   \textbf{I'm going to dice a little bit of peppers here.}                \\
        \midrule
            Skeleton        &   cooking                                                                 \\
                            &   chop yellow peppers                                                     \\ \cline{2-2}
            EDN             &   cook with a little pepper                                               \\
                            &   chop it little bit and sprinkle                                         \\
        \bottomrule     
    \end{tabular}
}
\end{table}

\noindent\textbf{How can computer vision do better?}
Our results show that the EDN model used as an out-of-the-box approach is not enough for sign language video generation. Specifically, we show that the model struggles with generating the hands and detailed facial expressions, which play a central role in sign language understanding.
We argue that human pose estimation plays an important key in this aspect and needs to be more robust to blurry images, especially in the hands and to fast movements in order to be suitable to sign language research.
We also argue that it is worth pursuing generative models that focus on generating hand details, particularly on the movements of the fingers, as well as clear facial expressions on full-body synthesis.


\label{sec:experiments}
\section{Conclusion}
In this paper, we present the \hts{} dataset, \emph{a large-scale multimodal and multiview dataset of American Sign Language}. With more than \textit{80 hours} of sign language videos and their corresponding speech signal, English transcripts and annotations,
\hts{} has the potential to impact a wide range of sign language understanding tasks, such as sign language recognition, translation and production, as well as wider multimodal and computer vision tasks like 3D human pose estimation. 
\hts{} extends the How2~\cite{How2} dataset, an existing multimodal dataset with a new sign language modality, and therefore enables connecting with research performed in the vision, speech and language communities.
In addition to that, we further conducted a study in which sign language videos generated from the automatically extracted annotations of our dataset were presented to ASL signers. To our knowledge, this is the first study how well keypoint-based synthetic videos, a commonly used representation of sign language production and translation, can be understood by sign language users. Our study indicates that current video synthesis methods allow the understanding to a certain extent \ie, the classification of the video category, but lack in fidelity to allow for a fine-grained understanding of the complete sign language sentence.

\label{sec:conclusion}
\section*{Acknowledgments}
\small{This work received funding from Facebook through gifts to CMU and UPC; through projects TEC2016-75976-R, TIN2015-65316-P, SEV-2015-0493 and PID2019-107255GB-C22 of the Spanish Government and 2017-SGR-1414 of Generalitat de Catalunya.
This work used XSEDE's “Bridges” system at the Pittsburgh Supercomputing Center (NSF award ACI-1445606).
Amanda Duarte has received support from la Caixa Foundation (ID 100010434) under the fellowship code LCF/BQ/IN18/11660029.
Shruti Palaskar was supported by the Facebook Fellowship program. The authors would like to thank Chinmay H\'{e}jmadi, Xabier Garcia and Brandon Taylor for their help during the data collection and processing and Yannis Kalantidis for his valuable feedback.
This work would not be possible without the collaboration and feedback from the signers and the Deaf community involved throughout the project.}


{\small
\bibliographystyle{ieee_fullname}
\bibliography{references}

\begin{thebibliography}{10}\itemsep=-1pt

\bibitem{deafstats}
World Health~Organization 2019.
\newblock Deafness and hearing loss.
\newblock
  \url{https://www.who.int/news-room/fact-sheets/detail/deafness-and-hearing-loss}.
\newblock Accessed: 2019-05-19.

\bibitem{BSL-1K}
Samuel Albanie, G{\"u}l Varol, Liliane Momeni, Triantafyllos Afouras, Joon~Son
  Chung, Neil Fox, and Andrew Zisserman.
\newblock Bsl-1k: Scaling up co-articulated sign language recognition using
  mouthing cues.
\newblock In {\em European Confernce on Computer Vision (ECCV)}, 2020.

\bibitem{sl-structure}
David~F Armstrong, William~C Stokoe, and Sherman~E Wilcox.
\newblock {\em Gesture and the nature of language}.
\newblock Cambridge University Press, 1995.

\bibitem{ASLLVD}
Vassilis Athitsos, Carol Neidle, Stan Sclaroff, Joan Nash, Alexandra Stefan,
  Quan Yuan, and Ashwin Thangali.
\newblock The american sign language lexicon video dataset.
\newblock In {\em CVPRW'08.}, pages 1--8. IEEE, 2008.

\bibitem{LDA}
David~M Blei, Andrew~Y Ng, and Michael~I Jordan.
\newblock Latent dirichlet allocation.
\newblock {\em Journal of machine Learning research}, 3(Jan):993--1022, 2003.

\bibitem{bragg2019sign}
Danielle Bragg, Oscar Koller, Mary Bellard, Larwan Berke, Patrick Boudreault,
  Annelies Braffort, Naomi Caselli, Matt Huenerfauth, Hernisa Kacorri, Tessa
  Verhoef, et~al.
\newblock Sign language recognition, generation, and translation: An
  interdisciplinary perspective.
\newblock In {\em The 21st International ACM SIGACCESS Conference on Computers
  and Accessibility}, pages 16--31, 2019.

\bibitem{sl-barriers}
Ruth Butler, Tracey Skelton, and Gill Valentine.
\newblock Language barriers: Exploring the worlds of the deaf.
\newblock {\em Disability Studies Quarterly}, 21(4), 2001.

\bibitem{camgoz2020sign}
Necati~Cihan Camgoz, Oscar Koller, Simon Hadfield, and Richard Bowden.
\newblock Sign language transformers: Joint end-to-end sign language
  recognition and translation.
\newblock In {\em Proceedings of the IEEE/CVF Conference on Computer Vision and
  Pattern Recognition}, pages 10023--10033, 2020.

\bibitem{openpose}
Zhe Cao, Gines Hidalgo, Tomas Simon, Shih-En Wei, and Yaser Sheikh.
\newblock Openpose: realtime multi-person 2d pose estimation using part
  affinity fields.
\newblock {\em arXiv preprint arXiv:1812.08008}, 2018.

\bibitem{ASL-Lex}
N Caselli, Z Sevcikova, A Cohen-Goldberg, and K Emmorey.
\newblock Asl-lex: A lexical database for asl.
\newblock {\em Behavior Research Methods}, 2016.

\bibitem{EDN}
Caroline Chan, Shiry Ginosar, Tinghui Zhou, and Alexei Efros.
\newblock Everybody dance now.
\newblock In {\em ICCV}, 2019.

\bibitem{SLTranslation}
Necati Cihan~Camgoz, Simon Hadfield, Oscar Koller, Hermann Ney, and Richard
  Bowden.
\newblock Neural sign language translation.
\newblock In {\em CVPR}, pages 7784--7793, 2018.

\bibitem{ELAN}
Onno Crasborn and Sloetjes Han.
\newblock Enhanced elan functionality for sign language corpora.
\newblock {\em Journal of deaf studies and deaf education}, 2008.

\bibitem{goldmann1992overcoming}
Warren~R Goldmann and James~R Mallory.
\newblock Overcoming communication barriers: communicating with deaf people.
\newblock 1992.

\bibitem{DGS-Korpus}
Thomas Hanke, Marc Schulder, Reiner Konrad, and Elena Jahn.
\newblock Extending the public dgs corpus in size and depth.
\newblock In {\em LREC2020 - Workshop on the Representation and Processing of
  Sign Languages}, pages 75--82, 2020.

\bibitem{video-based}
Jie Huang, Wengang Zhou, Qilin Zhang, Houqiang Li, and Weiping Li.
\newblock Video-based sign language recognition without temporal segmentation.
\newblock In {\em AAAI}, 2018.

\bibitem{panoptic}
Hanbyul Joo, Hao Liu, Lei Tan, Lin Gui, Bart Nabbe, Iain Matthews, Takeo
  Kanade, Shohei Nobuhara, and Yaser Sheikh.
\newblock Panoptic studio: A massively multiview system for social motion
  capture.
\newblock In {\em Proceedings of the IEEE International Conference on Computer
  Vision}, pages 3334--3342, 2015.

\bibitem{MS-ASL}
Hamid Reza~Vaezi Joze and Oscar Koller.
\newblock Ms-asl: A large-scale data set and benchmark for understanding
  american sign language.
\newblock {\em arXiv preprint arXiv:1812.01053}, 2018.

\bibitem{korean_SL}
Sang-Ki Ko, Chang~Jo Kim, Hyedong Jung, and Choongsang Cho.
\newblock Neural sign language translation based on human keypoint estimation.
\newblock {\em Applied Sciences}, 9(13), 2019.

\bibitem{gloss-informal}
Jolanta Lapiak.
\newblock Gloss: transcription symbols.
\newblock \url{https://www.handspeak.com/learn/index.php?id=3}.
\newblock Accessed: 2019-08-20.

\bibitem{WLASL}
Dongxu Li, Cristian Rodriguez, Xin Yu, and Hongdong Li.
\newblock Word-level deep sign language recognition from video: A new
  large-scale dataset and methods comparison.
\newblock In {\em The IEEE Winter Conference on Applications of Computer
  Vision}, pages 1459--1469, 2020.

\bibitem{gloss}
Scott~K Liddell et~al.
\newblock {\em Grammar, gesture, and meaning in American Sign Language}.
\newblock Cambridge University Press, 2003.

\bibitem{RVL-SLLL}
Aleix~M Mart{\'\i}nez, Ronnie~B Wilbur, Robin Shay, and Avinash~C Kak.
\newblock Purdue rvl-slll asl database for automatic recognition of american
  sign language.
\newblock In {\em Proceedings. Fourth IEEE International Conference on
  Multimodal Interfaces}, pages 167--172. IEEE, 2002.

\bibitem{NCSLGR}
Carol Neidle and Christian Vogler.
\newblock A new web interface to facilitate access to corpora: Development of
  the asllrp data access interface (dai).
\newblock In {\em Proc. 5th Workshop on the Representation and Processing of
  Sign Languages: Interactions between Corpus and Lexicon, LREC}, 2012.

\bibitem{papineni2002bleu}
Kishore Papineni, Salim Roukos, Todd Ward, and Wei-Jing Zhu.
\newblock Bleu: A method for automatic evaluation of machine translation.
\newblock In {\em ACL}, 2002.

\bibitem{reilly1980american}
Judy Reilly and Marina~L McIntire.
\newblock American sign language and pidgin sign english: What's the
  difference?
\newblock {\em Sign Language Studies}, pages 151--192, 1980.

\bibitem{How2}
Ramon Sanabria, Ozan Caglayan, Shruti Palaskar, Desmond Elliott, Lo{\"\i}c
  Barrault, Lucia Specia, and Florian Metze.
\newblock How2: a large-scale dataset for multimodal language understanding.
\newblock {\em arXiv preprint arXiv:1811.00347}, 2018.

\bibitem{saunders2020adversarial}
Ben Saunders, Necati~Cihan Camgoz, and Richard Bowden.
\newblock Adversarial training for multi-channel sign language production.
\newblock In {\em The 31st British Machine Vision Virtual Conference (BMVC)},
  2020.

\bibitem{saunders2020everybody}
Ben Saunders, Necati~Cihan Camgoz, and Richard Bowden.
\newblock Everybody sign now: Translating spoken language to photo realistic
  sign language video.
\newblock {\em arXiv preprint arXiv:2011.09846}, 2020.

\bibitem{progressive_transformers}
Ben Saunders, Necati~Cihan Camgoz, and Richard Bowden.
\newblock Progressive transformers for end-to-end sign language production.
\newblock In {\em European Confernce on Computer Vision (ECCV)}, 2020.

\bibitem{bsl-corpus}
Adam Schembri, Jordan Fenlon, Ramas Rentelis, Sally Reynolds, and Kearsy
  Cormier.
\newblock Building the british sign language corpus.
\newblock {\em Language Documentation \& Conservation}, 7:136--154, 2013.

\bibitem{ASLstructure}
William~C Stokoe~Jr.
\newblock Sign language structure: An outline of the visual communication
  systems of the american deaf.
\newblock {\em Journal of deaf studies and deaf education}, 10(1):3--37, 2005.

\bibitem{Text2Sign}
Stephanie Stoll, Necati~Cihan Camgoz, Simon Hadfield, and Richard Bowden.
\newblock Text2sign: towards sign language production using neural machine
  translation and generative adversarial networks.
\newblock In {\em International Journal of Computer Vision}, 2020.

\bibitem{signor}
{\v{S}}pela Vintar, Bo{\v{s}}tjan Jerko, and Marjetka Kulovec.
\newblock Compiling the slovene sign language corpus.
\newblock In {\em 5th Workshop on the Representation and Processing of Sign
  Languages: Interactions between Corpus and Lexicon. Language Resources and
  Evaluation Conference (LREC)}, volume~5, pages 159--162, 2012.

\bibitem{SIGNUM}
U. Von~Agris and K.-F. Kraiss.
\newblock Signum database: Video corpus for signer-independent continuous sign
  language recognition.
\newblock In {\em Workshop on Representation and Processing of Sign Languages},
  pages 243--246, 2010.

\bibitem{wang2018pix2pixHD}
Ting-Chun Wang, Ming-Yu Liu, Jun-Yan Zhu, Andrew Tao, Jan Kautz, and Bryan
  Catanzaro.
\newblock High-resolution image synthesis and semantic manipulation with
  conditional gans.
\newblock In {\em Proceedings of the IEEE Conference on Computer Vision and
  Pattern Recognition}, 2018.

\bibitem{woodward1972implications}
James~C Woodward.
\newblock Implications for sociolinguistic research among the deaf.
\newblock {\em Sign Language Studies}, pages 1--7, 1972.

\bibitem{PCK}
Yi Yang and Deva Ramanan.
\newblock Articulated human detection with flexible mixtures of parts.
\newblock {\em IEEE TPAMI}, 35:2878--90, 12 2013.

\bibitem{Boston104}
Morteza Zahedi, Philippe Dreuw, David Rybach, Thomas Deselaers, and Hermann
  Ney.
\newblock Continuous sign language recognition-approaches from speech
  recognition and available data resources.
\newblock In {\em Workshop on Representation and Processing of Sign Languages},
  2006.

\bibitem{WordsAreOurGlosses}
Jan Zelinka and Jakub Kanis.
\newblock Neural sign language synthesis: Words are our glosses.
\newblock In {\em The IEEE Winter Conference on Applications of Computer
  Vision}, pages 3395--3403, 2020.

\bibitem{Czech_Sign_Language}
Jan Zelinka, Jakub Kanis, and Petr Salajka.
\newblock Nn-based czech sign language synthesis.
\newblock In {\em International Conf. on Speech and Computer}, pages 559--568.
  Springer, 2019.

\end{thebibliography}
}

\newpage
\begin{center}
\twocolumn\textbf{\Large{Supplementary Material}}
\end{center}

\beginsupplement

\section{Sign Language}
In this section we discuss in more detail some important non-manual features (that are not conveyed through other linguistic parameters \eg palm orientation, handshape, etc.) present in sign languages.
It is important to remember that American Sign Language, for example, requires more than just complex hand movements to convey a message.  Without the use of proper facial expressions and other non-manual features as the ones described below, a message could be greatly misunderstood~\cite{ASLstructure}. 

\noindent\textbf{Head movement.} The movement of the head supports the semantics of sign language. Questions, affirmations, denials, and conditional clauses are communicated with the help of the signer’s head movement.

\noindent\textbf{Facial grammar.} Facial grammar does not only reflect a person’s affect and emotions, but also constitutes to large part of the grammar in sign languages. For example, a change of head pose combined with the lifting of the eye brows corresponds to a subjunctive.

\noindent\textbf{Mouth morphemes (mouthing).} 
Mouth movement or mouthing is used to convey an adjective, adverb, or another descriptive meaning in association with an ASL word.
Some ASL signs have a permanent mouth morpheme as part of their production. For example, the ASL word NOT-YET requires a mouth morpheme (TH) whereas LATE has no mouth morpheme. These two are the same sign but with a different non-manual signal.
These mouth morphemes are used in some contexts with some ASL signs, not all of them.

\section{How2Sign dataset}
Here we discuss some additional metadata that are important for a better understanding of our data as well as the biases and generalization of the systems trained using the How2Sign dataset. We also describe information that might be helpful for future similar data collection.

\begin{figure*}
    \centering
    \includegraphics[width=.9\textwidth]{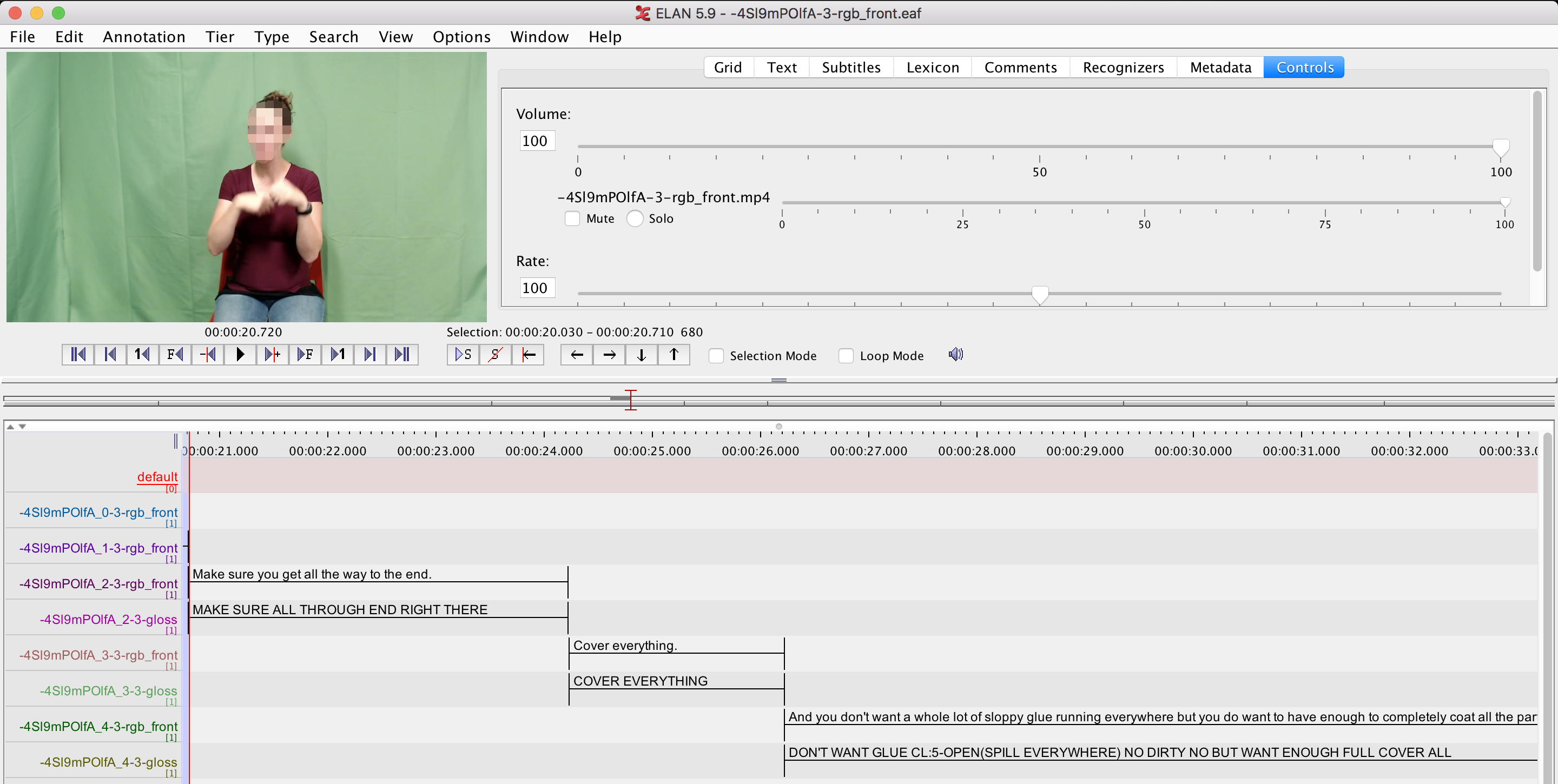}
    \caption{Samples of gloss annotations collected using ELAN.}
    \label{fig:ELAN-samples}
\end{figure*}

\noindent\textbf{Gloss.} We collected gloss annotations for the ASL videos present in the \hts{} dataset using ELAN. Figure \ref{fig:ELAN-samples} shows samples of the gloss annotations present in our dataset. Here we describe some conventional and few modified symbols and explanations that will be found in our dataset. A complete list is available on the dataset website.

\begin{itemize}
    \item\textit{Capital letters.} English glosses are written using capital letters. They represent an ASL word or sign. It is important to remember that gloss is not a translation. It is only an approximate representation of the ASL sign itself, not necessarily a meaning.
    \item A \textit{hyphen} is used to represent a single sign when more than one English word is used in gloss (\eg STARE-AT).
    \item The \textit{plus sign (+)} is used in ASL compound words (\eg MOTHER+FATHER -- used to transcribe parents). It is also used when someone combines two signs in one (\eg YOU THERE will be glossed as YOU+THERE).
    \item The \textit{plus sign (++)} at the end of a gloss indicates a number of repetitions of an ASL sign (\eg AGAIN++ -- the word ``again" was signed two more times meaning ``again and again").
    \item\textit{FS:} represents a fingerspelled word (\eg FS:AMELIA).
    \item\textit{IX} is a shortcut for ``index", which means to point to a certain location, object, or person.
    \item\textit{LOC} is a shortcut for ``locative", a part of the grammatical structure in ASL.
    \item\textit{CL:} is a shortcut for ``classifier". Classifiers are signs that use handshapes that are associated with specific categories (classes) of things, size, shape, or usage. They can help to clarify the message, highlight specific details, and provide an efficient way of conveying information\footnote{More info about handshapes and classifiers can be found at: \url{https://www.lifeprint.com/asl101/pages-signs/classifiers/classifiers-main.htm}}. 
    In our annotations, classifiers will appear as: ``CL:classifier(information)". For example, if the signer signs “TODAY BIKE" and uses a classifier to show the bike going up the hill, this would be glossed as: ``TODAY BIKE CL:3 (going uphill)").
\end{itemize}

\begin{figure*}
    \centering
    \includegraphics[width=.9\textwidth]{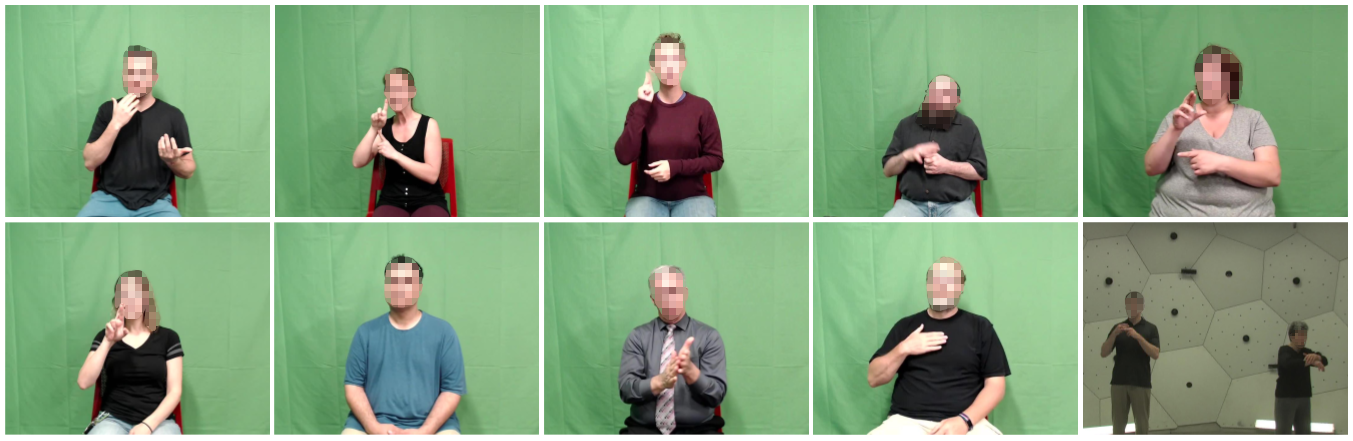}
    \caption{All the 11 signers that appear in the \hts{} dataset videos. On the top row, we can see signers 1-5 (from left to right) in the Green Screen Studio, while on the bottom row we can see signers 8-11 (again left to right) in the Green Screen Studio. The rightmost figure on the bottom row shows signers 6-7 in the Panoptic studio.}
    \label{fig:all_signers}
\end{figure*}

\noindent\textbf{Signers.} Figure~\ref{fig:all_signers} show all the 11 signers that participated in the recordings of the \hts{} dataset. From the 11 signers, four of them (signers 1, 2, 3 and 10 ) participated in both the Green Screen studio and the Panoptic studio recordings. Signers 6 and 7 participated only in the Panoptic studio recordings, while signers 4, 5, 8, 9 and 11 participated only in the Green Screen recordings. The signer ID information of each video is also made available.

\noindent\textbf{Recording pipeline.}
\textit{Importance of providing the speech and original video to the signer before the recordings:} As part of the design phase of our data collection, signers were asked to perform English to ASL translation when given: (1) just text without reading it beforehand; (2) the video and text together but without seeing it previously and (3) text and video together and allowing them to watch it before the recording. The conclusions for each case were: (1) signers found it hard to understand and follow the lines at the same time, causing lots of pauses and confusion; (2) signers found it easier to understand and translate but still with some pauses and (3) the understanding and flow improved. 

\subsection{Discussion}

\noindent\textbf{How high is the quality of the extracted keypoints?} We conducted a number of studies to estimate the quality of the automatically extracted 2D poses. A number of sanity checks showed us that extracting keypoints in higher resolution (1280 x 720) resulted to pose estimation that have on average higher confidence --  53.4\% average keypoint confidence for high resolution versus 42.4\% confidence for low resolution (210 x 260). 
This difference is more prominent when different parts of the body are analyzed. Table~\ref{tab:confidence_score} show the different average confidence scores when OpenPose is extracted using high and low resolution videos. We see that both hands are the most harm when low resolution is used.  

More importantly, in Section 4 we present a study with native speakers and verified that our 2D keypoints are sufficient to a certain degree for sign language users to classify and transcribe the ASL videos back to English.

\begin{table}
\caption{Average of confidence score of OpenPose on high resolution (1280 x 720) compared with low resolution (210 x 260) videos of the \hts{} dataset.} 
\label{tab:confidence_score}
\resizebox{\columnwidth}{!}{
\begin{tabular}{lcccc|c}
\toprule
                    & Body & Right hand & Left hand & Face & Total \\ 
\midrule
    High resolution & 0.39 & 0.42       & 0.47      & 0.84 & 0.53  \\ 
\midrule
    Low resolution  & 0.40 & 0.24       & 0.30      & 0.73 & 0.42  \\ 
\bottomrule
\end{tabular}}
\end{table}

\noindent\textbf{Factors that may impair accurate automatic tracking.} During the recording, signers were requested to not use loose clothes, rings, earrings, watch, or any other accessories that might impair accurate automatic tracking. They were also asked to wear solid colored shirts (that contrast with their skin tone).

\noindent\textbf{Out-of-vocabulary and signer generalization.} Although not specifically designed for this, the \hts{} dataset can be used for measuring generalization with respect to both out-of-vocabulary words and signers. The dataset contains 413 and 510 out-of-vocabulary words, \eg words that occur in validation and test, respectively, but not in training. It further contains duplicate recordings on the test set by a signer that is \textit{not present in the training set}; these recordings can be used for measuring generalization across different signers and help understand how well the models can recognise or translate the signs given an out of the distribution subject. 

\begin{figure*}[]
    \centering
    \includegraphics[width=.9\textwidth]{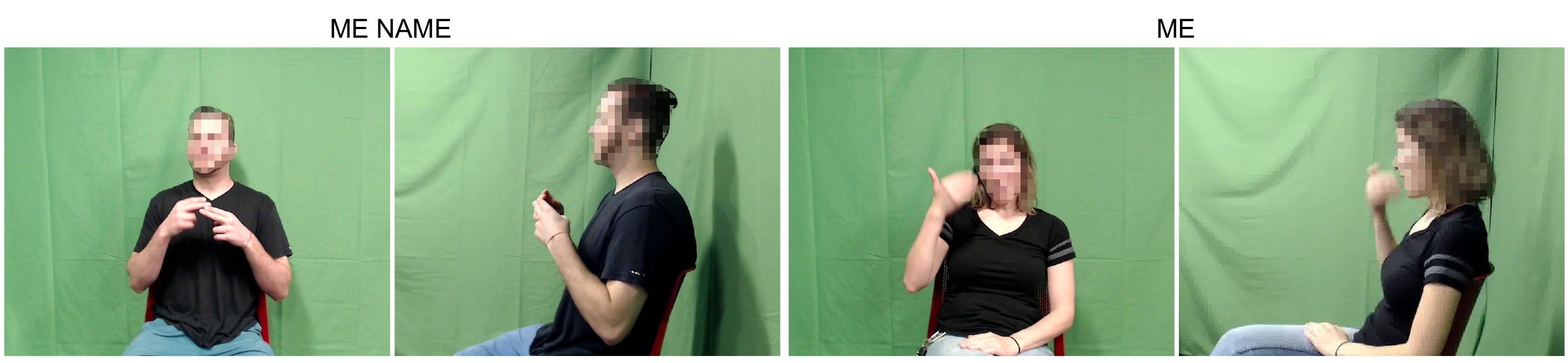}
    \caption{Sample of language variety on our dataset. Both signers were translating the sentence ``I am". We can see that the signer on the left used the casual approach of signing it (ME NAME) while the signer on the left used the formal approach (ME). }
    \label{fig:diff_sign}
\end{figure*}

\noindent\textbf{Language variety.} As discussed in subsection 3.5 our dataset contains variations in the language used during the recordings by each signer.
In addition to that, we also would like to mention that sign language speakers can also use different signs or different linguistic registers (\ie, formal or casual) to express the same given sentence. As we can see in Figure ~\ref{fig:diff_sign}, two signers from our dataset used two different signs in a linguistic register to express the phrase ``I am". The signer on the left used the casual approach of signing (ME NAME) while the signer on the left used the formal approach (ME). 

\begin{figure*}
    \centering
    \includegraphics[width=.9\textwidth]{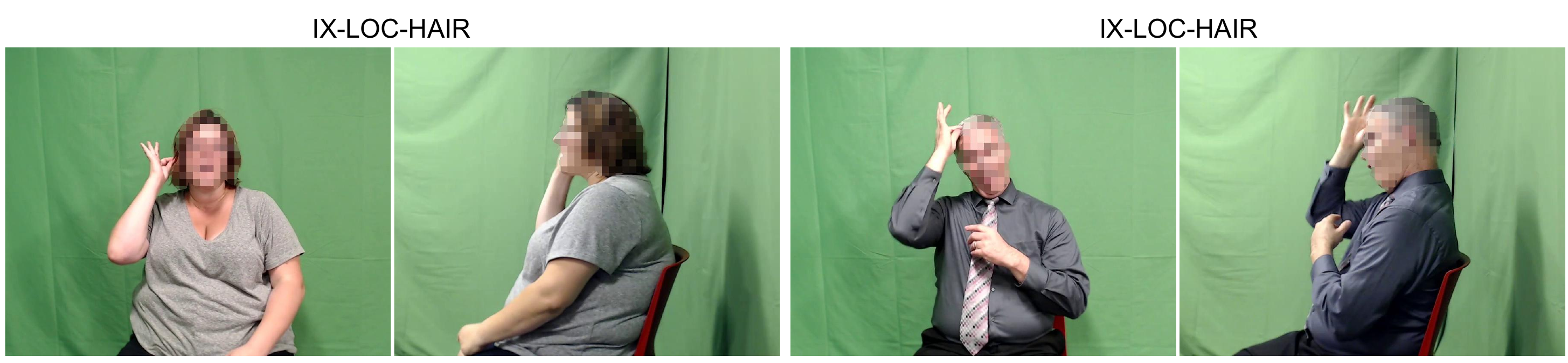}
    \caption{Sample of intra-sign variety. In this case, both signers are signing the word ``hair" (IX-LOC-HAIR). We can see that the on the left choose to point to her hair on a different position from the signer on the right.}
    \label{fig:diff_loc}
\end{figure*}

\noindent\textbf{Intra-sign variety.} In addition to the variety of signs and linguistic registers, it is also common to notice differences in the way of performing the same sign. For example, we can see on Figure~\ref{fig:diff_loc} two signers from our dataset signing the word ``hair". In this sign, as described by its gloss annotation (IX-LOC-HAIR) the signer points to their own hair location. While performing the sign, the person can use slightly different locations to point at. 

\subsection{How2Sign statistics per signer}
Table \ref{tab:data_stats_signer} presents detailed statistics of the videos from the \hts{} dataset recorded in the \textit{Green Screen studio} grouped by signer.
\begin{table*}[]
\centering
\caption{Statistics of the \textit{Green Screen studio} data by signer. We present the number of videos recorded by signer (videos), together with the total duration of the recorded videos in hours (Hours) and the number of utterances (Utterances) of each signer. }
\label{tab:data_stats_signer}
\resizebox{\textwidth}{!}{\begin{tabular}{llllllllll|l} 
    \toprule
                & \textbf{Signer 1} & \textbf{Signer 2} & \textbf{Signer 3} & \textbf{Signer 4} & \textbf{Signer 5} & \textbf{Signer 8} & \textbf{Signer 9} & \textbf{Signer 10} & \textbf{Signer 11} & \textbf{Total}  \\ 
    \midrule
                & \multicolumn{9}{c}{\textbf{Train}} &   \\ 
    \cmidrule{2-11} 
        Videos  & 50                & 22                & 163               & 24                & 899               & 994               & 18                & -                  & 43                  & \textbf{2213}  \\
        Hours   & 1.89              & 0.82              & 3.80              & 0.82              & 31.59             & 28.28             & 0.67              & -                  & 1.72                & \textbf{69.59} \\ 
        Utterances & 892            & 422               & 1859              & 398               & 12102             & 14596             & 292               & -                  & 486                 & \textbf{31047} \\
    \midrule
                & \multicolumn{9}{c}{\textbf{Test}} &   \\ 
    \cmidrule{2-11}
        Videos  & 16                & 16                & 37                & -                 & 47                & 42                & -                 & 26                 & -                    & \textbf{184}  \\
        Hours   & 0.51              & 0.53              & 1.05              & -                 & 1.67              & 1.08              & -                 & 0.71               & -                    & \textbf{5.55} \\
        Utterances & 224            & 243               & 538               & -                 & 621               & 449               & -                 & 268                & -                    & \textbf{2343} \\
    \midrule
                & \multicolumn{9}{c}{\textbf{Validation}} &  \\ 
    \cmidrule{2-11} 
        Videos  & 17                & 19                & 27                & -                 & 37                & 32                & -                 & -                  & -                    & \textbf{132} \\
        Hours   & 0.57              & 0.68              & 0.65              & -                 & 1.20              & 0.79              & -                 & -                  & -                    & \textbf{3.89} \\
        Utterances & 276            & 270               & 306               & -                 & 454               & 433               & -                 & -                  & -                    & \textbf{1739}  \\
    \bottomrule
\end{tabular}}
\end{table*}

\end{document}